\documentclass[10pt,twocolumn,letterpaper]{article}

\usepackage{cvpr} 

\usepackage{tcolorbox}
\usepackage{colortbl}
\usepackage{diagbox}
\usepackage{cuted}
\usepackage{makecell}
\usepackage{multirow}
\usepackage{array}

\newcommand{\egno}{\textit{e}.\textit{g}.} 
\newcommand{\ieno}{\textit{i}.\textit{e}.} 
\newcommand{\etcno}{\textit{etc}} 
\newcommand{\ours}{WeGen\xspace}
\newcommand{\ourdatasetshort}{DIIC\xspace}
\newcommand{\ourdatasetlong}{Dynamic Instance Identity Consistency\xspace}
\newcommand{\ourtarget}{user-friendly design copilot\xspace}
\newcommand{\ourrewritelong}{Prompt Self-Rewriting\xspace}
\newcommand{\ourrewriteshort}{PSR\xspace}

\renewcommand{\thefootnote}{\fnsymbol{footnote}}

\definecolor{cvprblue}{rgb}{0.21,0.49,0.74}
\usepackage[pagebackref,breaklinks,colorlinks,allcolors=cvprblue]{hyperref}

\def\paperID{6825} 
\def\confName{CVPR}
\def\confYear{2025}

\title{WeGen: A Unified Model for Interactive Multimodal Generation as We Chat}

\author{Zhipeng Huang\textsuperscript{1$\ddagger$}\footnotemark[1] \quad Shaobin
Zhuang\textsuperscript{2$\ddagger$}\footnotemark[1] \quad Canmiao Fu\textsuperscript{3}
\quad Binxin Yang\textsuperscript{3} \quad Ying Zhang\textsuperscript{3} \quad
Chong Sun\textsuperscript{3} \\ \quad Zhizheng Zhang\textsuperscript{5}\footnotemark[2]
\quad Yali Wang\textsuperscript{4}\footnotemark[2] \quad Chen Li\textsuperscript{3}
\quad Zheng-Jun Zha\textsuperscript{1} \\ \textsuperscript{\rm 1}University of
Science and Technology of China \quad \textsuperscript{\rm 2}Shanghai Jiao Tong
University \\ \textsuperscript{\rm 3}WeChat Vision, Tencent Inc. \quad \textsuperscript{\rm 4}Chinese
Academy of Sciences \quad \textsuperscript{\rm 5}Galbot }

\begin{document}
    \maketitle

    \renewcommand{\thefootnote}{\fnsymbol{footnote}}
    \footnotetext[1]{Equal contribution \textsuperscript{$\ddagger$}Work done as
    interns at WeChat} \footnotetext[2]{Corresponding authors (zhangzz@galbot.com, yl.wang@siat.ac.cn)}
    \renewcommand{\thefootnote}{\arabic{footnote}}

    \begin{strip}
        \vspace{-1.2cm}
        \hspace*{-0.85cm}
        \centering
        \includegraphics[width=1.09\textwidth]{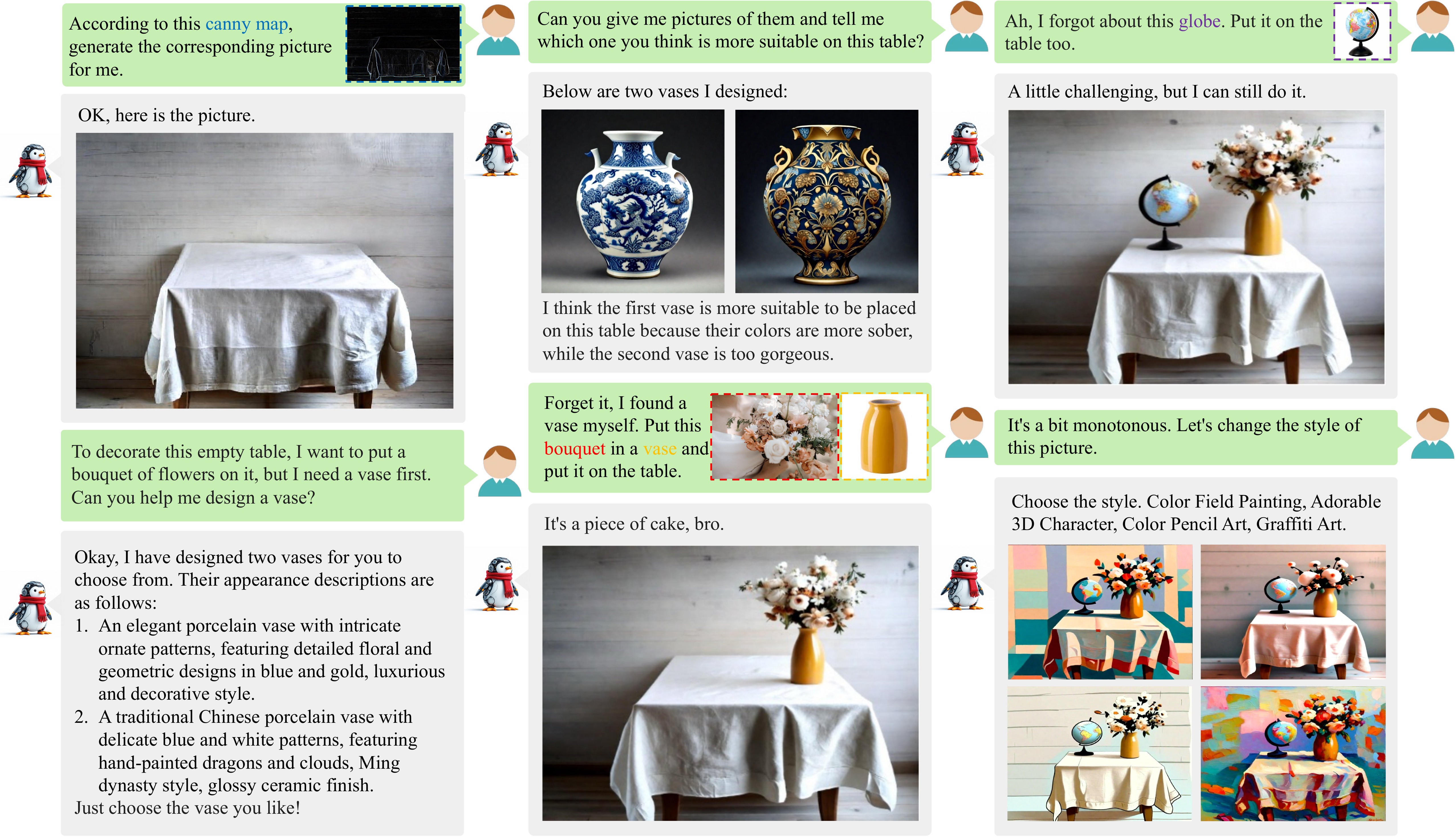}
        \captionsetup{type=figure} \captionof{figure}{Interactive dialogue examples between users and \ours, demonstrating unified capabilities across diverse visual generation tasks through natural conversations.}
        \label{fig:overview_dialog}
    \end{strip}


    \begin{abstract}
        Existing multimodal generative models fall short as qualified design copilots,
        as they often struggle to generate imaginative outputs once instructions
        are less detailed or lack the ability to maintain consistency with the
        provided references. In this work, we introduce \ours, a model that unifies
        multimodal generation and understanding, and promotes their interplay in
        iterative generation. It can generate diverse results with high creativity
        for less detailed instructions. And it can progressively refine prior
        generation results or integrating specific contents from references
        following the instructions in its chat with users. During this process,
        it is capable of preserving consistency in the parts that the user is
        already satisfied with. To this end, we curate a large-scale dataset, extracted
        from Internet videos, containing rich object dynamics and auto-labeled
        dynamics descriptions by advanced foundation models to date. These two information
        are interleaved into a single sequence to enable \ours to learn consistency-aware
        generation where the specified dynamics are generated while the consistency
        of unspecified content is preserved aligned with instructions. Besides, we
        introduce a prompt self-rewriting mechanism to enhance generation diversity.
        Extensive experiments demonstrate the effectiveness of unifying
        multimodal understanding and generation in \ours and show it achieves
        state-of-the-art performance across various visual generation benchmarks.
        These also demonstrate the potential of \ours as a user-friendly design copilot
        as desired. The code and models will be available at
        \url{https://github.com/hzphzp/WeGen}.
    \end{abstract}



    \section{Introduction}
    \label{sec:intro}

    Recent years have witnessed remarkable progress in AI-powered visual generation
    technologies, marked by numerous groundbreaking models like Stable Diffusion~\cite{rombach2022high}
    and its variants~\cite{zhang2023adding,ye2023ip,ruiz2022dreambooth}.
    However, the practical application of these tools remains challenging for
    general users\textemdash each visual task typically requires a dedicated model,
    and users need to organize multiple specialized component models and
    design the workflow carefully.
    Unlike ChatGPT's intuitive interface that has enabled widespread adoption,
    current visual generation tools remain challenging to use due to their steep
    learning curves. This motivates \ourtarget, a system that enables natural
    multimodal interaction with its inherent diverse generation capabilities (Figure~\ref{fig:overview_dialog}).

    To address these requirements, we propose \ours. As shown in Fig.~\ref{fig:overview_dialog},~\ref{fig:case_study},
    and~\ref{fig:grounding_result1}, \ours seamlessly integrates diverse
    capabilities including multimodal understanding, text-to-image generation,
    subject-driven generation with identity preservation, condition-driven generation,
    image restoration, and style transfer, \etcno. It enables users to achieve various
    visual generation goals through natural conversation, eliminating the
    complexity of carefully orchestrating multiple task-specific models.

    \ours combines Multimodal Large Language Models (MLLM) and diffusion model
    to achieve this versatility. The MLLM component, built upon CLIP~\cite{radford2021learning,sun2023eva}
    and LLM~\cite{touvron2023llama} architectures, enables natural dialogue interactions.
    Meanwhile, the diffusion model backbone ensures high-quality visual generation.
    Crucially, we leverage the MLLM's capabilities to standardize the injection of
    various visual and language conditions, allowing \ours to unify multiple
    textual and visual generation tasks under a single framework.

    While this combination of MLLMs and diffusion models presents a promising
    direction for unified modeling, recent preliminary explorations~\cite{ge2023planting,ge2023making,sun2023emu,ge2024seed,sun2024generative}
    have revealed some fundamental challenges that need to be addressed, which
    can be summarized into two issues.

    First, maintaining instance identity consistency with user-provided reference
    images remains challenging, yet crucial for practical applications (As shown
    in Figure~\ref{fig:overview_dialog}, where \ourtarget should preserve the
    user-selected vase across generations). Users often need to selectively
    retain key attributes from reference images, such as faces, landmark
    buildings, \etcno, while allowing reasonable variations in other aspects (pose,
    expression, lightning, \etcno). Simply copy-paste the entire reference image
    is not desirable, as it limits creativity and the rationality of the whole picture.
    It is crucial to balance preserved key features with natural variation. Second,
    generating diverse outputs from a single instruction remains challenging for
    existing methods (Fig.~\ref{fig:random_output}), As shown in Figure~\ref{fig:overview_dialog},
    when users have only vague initial preferences (e.g., "a vase" without
    specific details), \ourtarget should offer diverse alternatives for
    selection. However, previous methods tend to produce similar outputs even
    with different random seeds, as they directly map condition to continuous visual
    features for diffusion models without sampling process of discrete visual token.
    This deterministic mapping lacks the natural diversity, while attempts to discretize
    CLIP features for sampling lead to significant information loss.

    To tackle the instance identity consistency challenge, as shown in Fig.~\ref{fig:ablation_vis},
    we explore the scaling law when CLIP is used as an encoder
    and introduce the \ourdatasetlong (\ourdatasetshort) data pipeline by
    tracking entities across video sequences and capturing how they naturally
    while maintaining identity.
    To enhance generation diversity, we introduce an \ourrewritelong (\ourrewriteshort)
    that leverages MLLM's language capabilities to rewrite a detail image
    description before generate image features. \ourrewriteshort introduces randomness
    through additional discrete text token sampling, allowing the model to
    explore different interpretations while maintaining semantic alignment with
    instructions.

    In summary, the contributions of this paper can be summarized in the following
    four points:

    \begin{itemize}
        \item We propose \ours, a unified framework that serves as a \ourtarget
            by integrating diverse textual and visual generation capabilities
            into a single model with natural conversation interface, making advanced
            visual generation accessible to general users.

        \item We introduce the \ourdatasetlong (\ourdatasetshort) data pipeline to
            tackle the instance identity consistency challenge and balance preserved
            key features with natural variation.

        \item We propose \ourrewritelong (\ourrewriteshort) to enhance generation
            diversity, which introduces randomness through discrete text token
            sampling, allowing the model to explore different interpretations while
            maintaining semantic alignment.

        \item Extensive experiments demonstrate that \ours achieves state-of-the-art
            performance across multiple visual generation benchmarks,
            particularly excelling in maintaining instance identity consistency
            while enabling creative variations.
    \end{itemize}

    \section{Related Work}
    \label{sec:related_work}

    \noindent
    \textbf{Multimodal Understanding Models.} Recent advancements in large language
    models (LLMs)~\cite{achiam2023gpt} have revealed their exceptional capabilities
    in understanding and generating text. To extend such capabilities beyond
    text, we have seen an emergence of Multimodal Large Language Models (MLLMs)~\cite{zhu2023minigpt,li2023blip,lu2024deepseek}.
    These works essentially endow the LLMs with multimodal capabilities by
    aligning the visual encoder with the features of the LLMs. MiniGPT-4~\cite{zhu2023minigpt}
    and LLaVA~\cite{liu2023visualinstructiontuning} align a frozen visual encoder
    with the advanced LLM Vicuna~\cite{Zheng2023JudgingLW} using a single projection
    layer, exhibiting abilities similar to GPT-4~\cite{openai2024gpt4technicalreport}.
    BLIP-2~\cite{li2023blip} introduces a lightweight Querying Transformer that effectively
    bridges the modality gap between the frozen image encoder and LLM through a two-stage
    training strategy. Although these works enable LLM to achieve multimodal
    understanding capabilities, they cannot extend LLM's generative capabilities
    from text to the visual domain.

    \noindent
    \textbf{Diffusion Models.} Diffusion models have achieved notable progress
    in synthesizing unprecedented high-quality images~\cite{nichol2021glide,ramesh2022hierarchical,saharia2022photorealistic,rombach2022high,Balaji2022eDiffITD,Kusupati2022MatryoshkaRL,Dai2023EmuEI,BetkerImprovingIG}
    and videos~\cite{singer2023makeavideo, blattmann2023align,
    Wang2023ModelScopeTT, Zhang2023Show1MP, Wang2023LAVIEHV, Ge2023PreserveYO,
    zeng2024make, chen2023seineshorttolongvideodiffusion, zhuang2024vlogger}. Typically,
    these methods encode the input into a continuous latent space with VAE~\cite{Esser2020TamingTF}
    and learn to model the denoising diffusion process. This framework injects condition
    into the diffusion model through cross attention to generate desired results
    aligned with the condition. However, extending these base models to specific
    generation tasks requires task-specific model architecture design~\cite{zhang2023adding,ye2023ip,ruiz2023dreambooth,hu2021lora}
    and pipeline engineering~\cite{comfyui2023}.

    \noindent
    \textbf{Unified Models for Multimodal Understanding and Visual Generation.}
    Recent works have explored combining MLLMs with diffusion models to create
    unified frameworks for multimodal understanding and generation. Starting with
    GILL~\cite{koh2024generating}, followed by Emu~\cite{sun2023emu} and SEED-LLaMA~\cite{ge2023making}
    these approaches aim to leverage MLLMs' understanding capabilities alongside
    diffusion models' generation power. Recent attempts like SEED-X~\cite{ge2024seed}
    maintain instance identity consistency by directly using reference images as
    latent maps concatenated with noise input of diffusion decoder.
    While this ensures strong visual similarity of the origin image, it also limits
    the decoder's applicability in many visual generation tasks. Furthermore, while
    methods like Emu2~\cite{sun2024generative} separate understanding and generation
    into distinct models with limited generation capabilities (only text-to-image
    and subject-drive generation), our approach integrates both aspects into a
    single unified model supporting a wide range of tasks. Another class of visual
    generative models is based on auto-regressive models~\cite{tian2024visual,team2024chameleon,liu2024lumina,kondratyuk2023videopoet}.
    These methods encode inputs and outputs into discrete space via VQVAE~\cite{Esser2020TamingTF}
    and then model the generation process by the paradigm of next-token
    prediction.
    However, these methods typically require prohibitive computational costs to
    bridge the gap between text and image modalities in order to achieve
    comparable results to the aforementioned approaches.

    \begin{figure*}[t]
        \centering
        \includegraphics[width=0.9\linewidth]{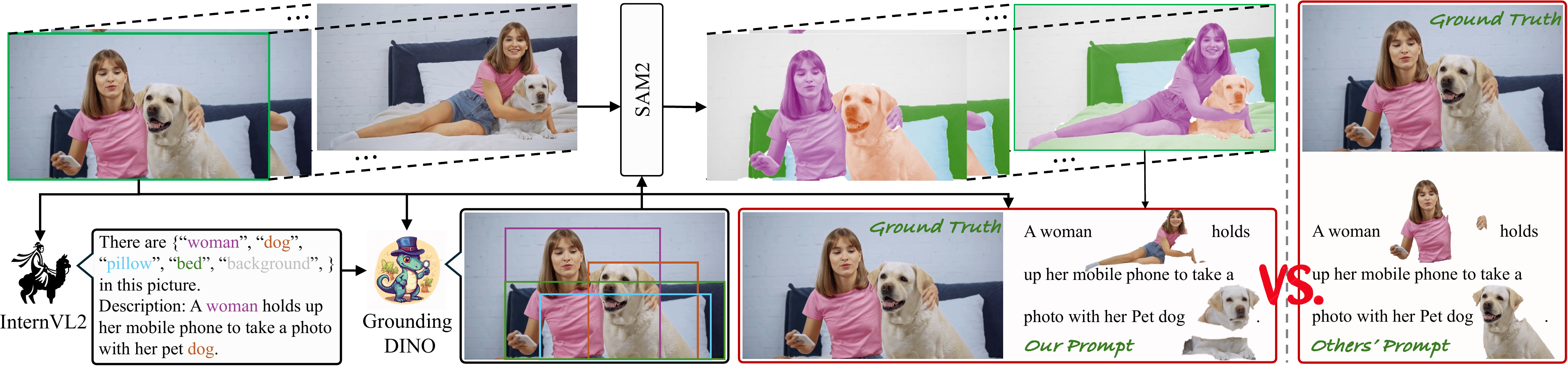}
        \vspace{-0.3cm}
        \caption{ \textbf{\ourdatasetlong (\ourdatasetshort) Data-pipeline.} }
        \label{fig:data_pipe}
        \vspace{-0.3cm}
    \end{figure*}

    \section{Method}
    \label{sec:method}
    In this section, we present \ours, a unified framework for multimodal
    understanding and visual generation with diversities.
    We first present the overall architecture of \ours, followed by training
    pipeline(\S\ref{subsec:overall}).
    We then address two key challenges: maintaining instance identity
    consistency through our \ourdatasetlong (\ourdatasetshort) data pipeline (\S\ref{subsec:tdic}),
    and enhancing generation diversity via \ourrewritelong (\ourrewriteshort)
    mechanism (\S\ref{subsec:method_diversity}).

    \subsection{Overall}
    \label{subsec:overall}

    \noindent
    \textbf{Architecture.} Following previous works~\cite{koh2024generating,pan2023kosmos,sun2024generative},
    our model is composed of three primary components:
    a CLIP encoder that transforms reference images into 64 visual features,
    a large language model (LLM) that processes alternating text and visual
    inputs and generates multi-modal output embeddings,
    and an SDXL~\cite{podell2023sdxl} decoder that converts the generated features
    into the final image.

    \noindent
    \textbf{Training.} We employ a two-stage training pipeline: First, we train
    the SDXL decoder to reconstruct images from CLIP-encoded features.
    The CLIP encoder remains frozen while SDXL is fully fine-tuned with diffusion
    loss.
    Second, we conduct LLM training with interleaved visual-text data.
    Keeping model weights of both CLIP and SDXL frozen, we fine-tune the LLM on various
    tasks including understanding, text-to-image generation, editing, and
    conditional generation.
    All tasks are reformulated into our interleaved format, with text tokens and
    visual features supervised by category and regression loss, respectively.

    \subsection{Dynamic Instance Identity Consistency}
    \label{subsec:tdic}

    In subject-driven generation tasks, maintaining instance identity consistency
    (\emph{i.e.}, preserving essential instance attributes from reference images
    while allowing natural variations) is a key challenge.
    This challenge stems from two limitations in current approaches~\cite{ge2023planting,ge2023making,sun2023emu,ge2024seed}:
    1) Information loss during encoding-decoding: Existing methods struggle to accurately
    reconstruct input images (see supplementary materials), leading to degraded
    instance recognition features.
    2) Limited training data: Using single-image for training provides only identical
    input-output pairs, encouraging simple copy-paste behavior rather than
    learning meaningful attribute preservation editing with moderate appearance
    changes.
    To address these issues, we propose two solutions: adopting a large-scale CLIP
    encoder to minimize information loss (Figure~\ref{fig:ablation_vis}(a)), and
    introducing the \ourdatasetshort data pipeline that leverages video sequences
    to capture natural instance variations while maintaining identity consistency
    (\S\ref{subsec:tdic}).

    As shonw in Figure~\ref{fig:data_pipe}, previous works~\cite{sun2024generative,
    pan2023kosmos} train on single-image segmentation datasets where input and output
    instances are exactly identical, leading to simple copy-paste behavior. While
    they attempt to introduce variations through artificial augmentations (flipping,
    color jittering), these synthetic changes fail to capture natural instance variations
    in pose, expression, and context, often resulting in unrealistic artifacts. Our
    \ourdatasetshort data pipeline constructed from videos, where instances
    naturally changes through time while maintaining their core identity.

    As shown in Fig.~\ref{fig:data_pipe}, we collect videos from various
    platforms~\cite{nan2024openvid,chen2024panda,wang2023internvid} and apply a
    three-step filtering process: (1) removing videos with subtitles using OCR~\cite{easyocr},
    (2) filtering out videos with abrupt scene changes via motion detection~\cite{xu2023unifying},
    and (3) selecting high-quality videos through aesthetic assessment~\cite{wu2023exploring}.
    This process retains approximately 70\% of the original videos for training.
    Our annotation process consists of four key steps for each filtered video:

    \noindent
    \textbf{Instance Identification}: Given a video sequence
    $\{x_{t}\}_{t=1}^{T}$, we first select the initial frame $x_{1}$ (Fig.~\ref{fig:data_pipe}).
    We prompt InternVL2~\cite{chen2023internvl,chen2024far} to generate precise captions
    (\egno, ``A girl in pink shirt and golden hair is playing with her golden retriever
    on the bed") and extract noun chunks $\{n_{i}\}_{i=1}^{N}$ to identify instances
    (\egno, ``girl", ``golden retriever", ``bed"). This approach, rather than directly
    parsing existing captions with tools like spaCy~\cite{honnibal2020industrial,peng2023kosmos},
    avoids abstract nouns (\egno, ``time", ``love", ``freedom") that are
    difficult to visually ground. Moreover, by generating captions on-the-fly, our
    method generalizes to any video sequence without requiring pre-existing annotations,
    significantly expanding its applicability.

    \noindent
    \textbf{Bounding Box Detection}: For each identified instance $i$ (e.g., ``girl",
    ``golden retriever", "bed"), we apply Grounding DINO~\cite{zhang2022dino} to
    obtain precise bounding boxes $\{bbox_{i}\}_{i=1}^{N}$. Grounding DINO's
    zero-shot object detection capability ensures accurate localization even for
    instances not seen during training, making our pipeline robust to diverse object
    categories and scenes.

    \noindent
    \textbf{Instance Tracking}: For each instance $i$, we use its bounding box
    and box center as the prompt to initialize SAM2~\cite{ravi2024sam}, which generates
    instance segments $\{seg_{i}^{t}\}_{t=1}^{T}$ across the video sequence.
    This tracking process maintains consistent instance segmentation while
    adapting to natural pose and appearance variations over time.

    \noindent
    \textbf{Frame Pair Selection}: Given the tracked instances, we sample frame pairs
    with temporal interval $t_{ref}$ to construct our training data $\{(n_{i}, bb
    ox_{i}^{t_{ref}}, seg_{i}^{t_{ref}})_{i=1}^{N}, x_{1}\}$. For example, with $t
    _{ref}=25$, we capture how the girl's pose and the dog's position naturally
    vary while maintaining their identities and relationship. The interval $t_{ref}$
    controls the degree of variation - larger values capture more significant changes
    in pose and appearance, while smaller values focus on subtle frame-to-frame
    variations.

    \noindent
    \textbf{Description Construction}: We format training data into MLLM-compatible
    instruction format, where the context includes both caption and instance information
    (caption, noun chunks, bounding boxes and segmented images in $t_{ref}$).
    The structured format is shown below:
    \begin{tcolorbox}
        [colback=gray!5,boxrule=0pt] {\footnotesize\texttt{<p>A girl</p><b>$\{$$bbox_{1}^{t_{ref}}$$\}$</b><img>$\{$$seg_{1}^{t_{ref}}$$\}$\ </img> in pink shirt and golden hair is playing with her <p>golden retriever</p><b>$\{$$bbox_{2}^{t_{ref}}$$\}$</b><img>$\{$$seg_{2}^{t_{ref}}$$\}$\ </img> on the bed...}}
    \end{tcolorbox}
    \noindent
    {During training, bounding boxes and segmentation images are randomly dropped with 0.3 probability to enhance model robustness. The model learns to generate the first frame $x_{1}$ conditioned on the above structured input.}

    This approach enables the model to learn the balance between consistent and
    variable attributes. More details about our \ourdatasetshort dataset,
    including data statistics analysis, can be found in the supplementary material.
    We will release the dataset.

    \subsection{\ourrewritelong}
    \label{subsec:method_diversity}

    Current methods struggle to generate diverse outputs from the same prompt,
    producing nearly identical images despite different random seeds (Fig.~\ref{fig:random_output}).
    This limitation occurs because current methods generate images through
    deterministic regression of continuous visual features without an effective
    sampling mechanism, unlike the natural diversity in discrete text generation.
    To enhance generation diversity, we propose \ourrewriteshort, a \textit{plug-and-play}
    approach that requires no architectural changes which introduces controlled randomness
    through prompt rewriting before visual feature generation.

    \noindent
    \textbf{Training to learn \ourrewriteshort.} We leverage LAION~\cite{schuhmann2022laion},
    the large-scale open-source image dataset available. To avoid potential
    hallucinations from its noisy original captions, we generate new high-quality
    annotations using BLIP-2~\cite{li2023blip} and Qwen-VL~\cite{bai2023qwen} to
    obtain brief captions $c_{\text{brief}}$ and detailed descriptions $c_{\text{dense}}$
    respectively. For each image $x$, we construct instruction-tuning samples in
    the following format:
    \begin{tcolorbox}
        [colback=gray!5,boxrule=0pt] {\footnotesize \textit{User:} Generate an image with prompt rewrite about $\{$$c_{\text{brief}}$$\}$.

        \textit{Assistant:} Here is my detailed description: $\{$$c_{\text{dense}}$$\}$ Here is the generated image: \textless img\textgreater$\{$$x$$\}$\textless /img\textgreater. }
    \end{tcolorbox}
    \noindent
    {This approach enables the model to learn both prompt enhancement and image generation in a unified framework.}

    \noindent
    \textbf{Inference with \ourrewriteshort for diverse generation.} During
    inference, our model first enriches the input prompt through controlled sampling
    strategies (nucleus sampling and temperature sampling). The prompt rewriting
    process follows an autoregressive sampling procedure:
    \begin{equation}
        \text{P}(c_{\text{dense}}|c_{\text{brief}})=\prod \limits_{i=1}^{m}\text{P}
        (c_{d_{i}}|c_{\text{brief}}, c_{d_{1}}, \cdots ,c_{d_{i-1}}) \label{eq:rewrite}
    \end{equation}
    where $m$ is the length of generated detailed caption $c_{d}$, and
    randomness is introduced through sampling strategies during token generation.
    The complete generation process is:
    \begin{equation}
        \text{P}(c_{\text{dense}}, I|c_{\text{brief}})=\text{$\text{P}^{'}$}(I|c_{\text{brief}}
        , c_{\text{dense}})\text{P}(c_{\text{dense}}|c_{\text{brief}}) \label{eq:gen}
    \end{equation}
    \noindent
    {The generation diversity primarily comes from the second term $\text{P}(c_{\text{dense}}|c_{\text{brief}})$, where different sampling strategies create variations in the rewritten prompts, while $\text{$\text{P}^{'}$}(I|c_{\text{brief}}, c_{\text{dense}})$ is relatively deterministic.}
    This approach achieves two goals: enabling diverse outputs through sampling-based
    prompt rewriting while improving generation quality through enhanced prompt details~\cite{BetkerImprovingIG}.
    The semantic alignment is ensured by our training data, where both brief and
    detailed captions are high-quality descriptions of the same image, teaching
    the model to enrich details while staying faithful to the original content. We
    will release our rewriting dataset.


    \section{Experiments}
    \label{sec:exp}

    Our goal is to develop a unified framework for \ourtarget that handles multiple
    tasks with a single model. To validate our approach, we first demonstrate \ours's
    effectiveness across various visual generation tasks (\S\ref{subsec:multi_task}).
    We then evaluate our solutions to two key challenges: maintaining instance
    identity consistency (\S\ref{subsec:consistency}) and enabling generation diversity
    (\S\ref{subsec:exp_diversity}). Finally, through ablation studies (\S\ref{subsec:ablation}),
    we analyze how the \ourdatasetshort data pipeline, \ourrewriteshort mechanism
    contribute to these capabilities.

    \subsection{Implementation Details}
    \label{subsec:implementation} Following previous works~\cite{koh2024generating,pan2023kosmos,sun2024generative},
    we implement \ours using state-of-the-art components: EVA-CLIP~\cite{sun2023eva}
    as the visual encoder, SDXL as the diffusion decoder, and LLaMA-2-7B-chat~\cite{touvron2023llama}
    as the language model backbone. The model is trained on multiple carefully
    curated datasets (summarized in supplementary) using mixed precision (bfloat16)
    with DeepSpeed ZeRO-2 optimization across 64 A100 GPUs. Detailed training configurations
    and dataset statistics are provided in the supplementary materials.

    \subsection{Unified Multi-Task Generation}
    \label{subsec:multi_task} To evaluate \ours as a unified framework, 1) we first
    assess its text-to-image generation capability through quantitative metrics,
    as this forms the foundation for all visual generation tasks. 2) We then
    demonstrate the framework's versatility through comprehensive case studies
    across a wide range of tasks, Detailed quantitative evaluations for specific
    tasks and more case studies are provided in \S\ref{subsec:consistency} and
    the supplementary materials.

    \begin{table}[tbp]
        \centering
        \resizebox{1.0\linewidth}{!}{
        \begin{tabular}{@{}lcccc@{}}
            \toprule \textbf{Model}                     & \textbf{Params} & \textbf{BS $\times$ Iter} & \textbf{FID} (↓) & \textbf{CLIP-T} (↑) \\
            \midrule GLIDE~\cite{nichol2021glide}       & 3B              & 2048 $\times$ 2.5M        & 12.24            & --                  \\
            LDM~\cite{rombach2022high}                  & 1.45B           & 680 $\times$ 370K         & 12.63            & --                  \\
            Make-A-Scene~\cite{gafni2022make}           & 4B              & 1024 $\times$ 170K        & 11.84            & --                  \\
            DALL-E 2~\cite{ramesh2022hierarchical}      & 3.5B            & 2048 $\times$ 800K        & 10.39            & 0.314               \\
            SDv1.5~\cite{nichol2021glide}               & 0.8B            & 2048 $\times$ 860K        & 9.93             & 0.302               \\
            SDXL~\cite{podell2023sdxl}                  & 2.6B            & 2048 $\times$ 800K        & --               & 0.310               \\
            Imagen~\cite{saharia2022photorealistic}     & 2B              & 2048 $\times$ 2.5M        & 7.27             & 0.270               \\
            Ediff-I~\cite{balaji2022ediff}              & 9.1B            & 2048 $\times$ 800K        & 6.95             & --                  \\
            \midrule Chameleon~\cite{team2024chameleon} & 7B              & 488 $\times$ 250K         & 29.60            & 0.243               \\
            DALL-E~\cite{ramesh2021zero}                & 12B             & 1024 $\times$ 430K        & 27.50            & --                  \\
            MMAR~\cite{yang2024mmar}                    & 7B              & 1152 $\times$ 313K        & 17.10            & --                  \\
            SEED-X~\cite{ge2024seed}                    & 13B             & --                        & 12.68            & --                  \\
            GILL~\cite{koh2024generating}               & 6.7B            & 200 $\times$ 20K          & 12.20            & --                  \\
            Kosmos-G~\cite{pan2023kosmos}               & 1.9B            & 4688 $\times$ 300K        & 10.99            & -                   \\
            Emu~\cite{sun2023emu}                       & 13B             & 480 $\times$ 10K          & 11.66            & 0.286               \\
            Emu2-Gen~\cite{sun2024generative}           & 33B             & 6912 $\times$ 36K         & --               & 0.297               \\
            \rowcolor[gray]{0.9} \textbf{\ours(Ours)}   & 7B              & 2048 $\times$ 20K         & \textbf{9.39}    & \textbf{0.308}      \\
            \bottomrule
        \end{tabular}
        }
        \vspace{-0.3cm}
        \caption{Comparison of text-to-image generation methods on COCO2014
        dataset. We report model parameters (Params), training computation (BS $\times$
        Iter represents batch size $\times$ training iterations), and generation
        quality metrics (FID and CLIP-T score). Lower FID and higher CLIP-T
        scores indicate better performance.}
        \vspace{-0.3cm}
        \label{tab:stage2_gen}
    \end{table}

    \noindent
    \textbf{Text-to-Image Generation.} In the realm of text-to-image generation,
    there are two primary technical approaches: one relies on pure diffusion
    models, while the other leverages multimodal large language models (MLLMs).
    Our model falls into the latter category, where we have achieved state-of-the-art
    (SOTA) performance among MLLM-based approaches, achieving an FID score of
    9.39 and a CLIP-T score of 0.308, as shown in Table~\ref{tab:stage2_gen}.
    Notably, our model accomplishes this with less training data and reduced computational
    cost, while also supporting a wide range of tasks within a unified framework.
    Compared to diffusion-based models, our model's performance is comparable in
    terms of metrics, but it offers the advantage of supporting multiple tasks and
    possessing both language and visual understanding capabilities.

\begin{figure*}[!ht]
        \centering
        \includegraphics[width=\textwidth]{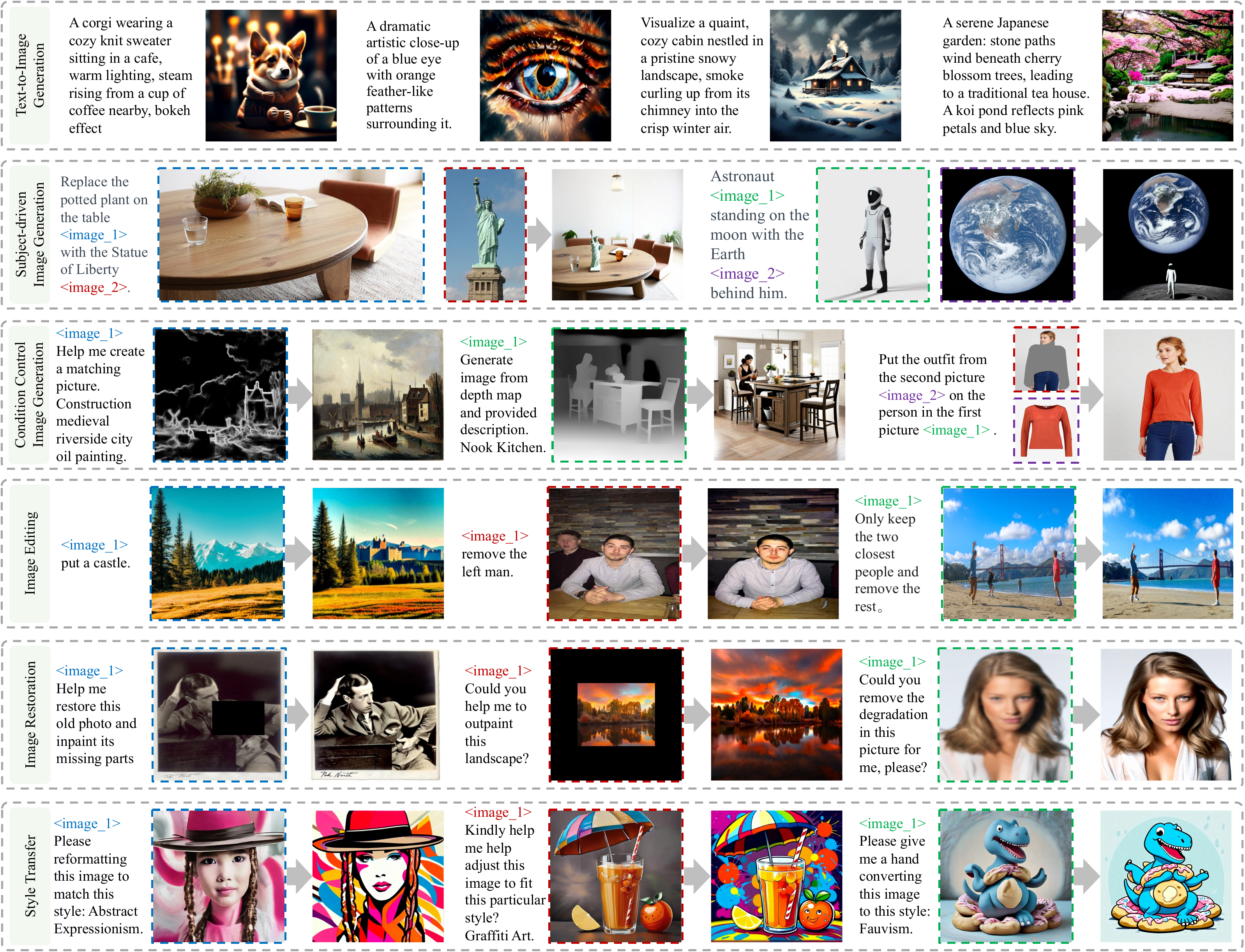}
        \vspace{-0.7cm}
        \caption{Case studies showcasing \ours's capabilities across various
        tasks, including text-to-image generation, subject-driven visual
        generation (both single and multiple subjects), image editing, condition-based
        generation (canny, depth, pose), style transfer, super-resolution, inpainting,
        outpainting}
        \label{fig:case_study}
        \vspace{-0.3cm}
    \end{figure*}

    \noindent
    \textbf{Case Studies on Diverse Tasks.} As shown in Figure~\ref{fig:case_study},
    we demonstrate \ours's capabilities across a wide range of tasks, including text-to-image
    generation, subject-driven generation, condition-based generation (canny, depth,
    pose), style transfer, super-resolution, inpainting, outpainting, and
    various editing operations. These qualitative results highlight our model's versatility
    across diverse visual generation tasks. Detailed quantitative evaluations
    for specific tasks can be found in \S\ref{subsec:consistency} and the supplementary
    materials, along with additional case studies.

    \subsection{Dynamic Instance Identity Consistency}
    \label{subsec:consistency}

\begin{figure*}[t]
        \centering
        \includegraphics[width=1\linewidth]{
            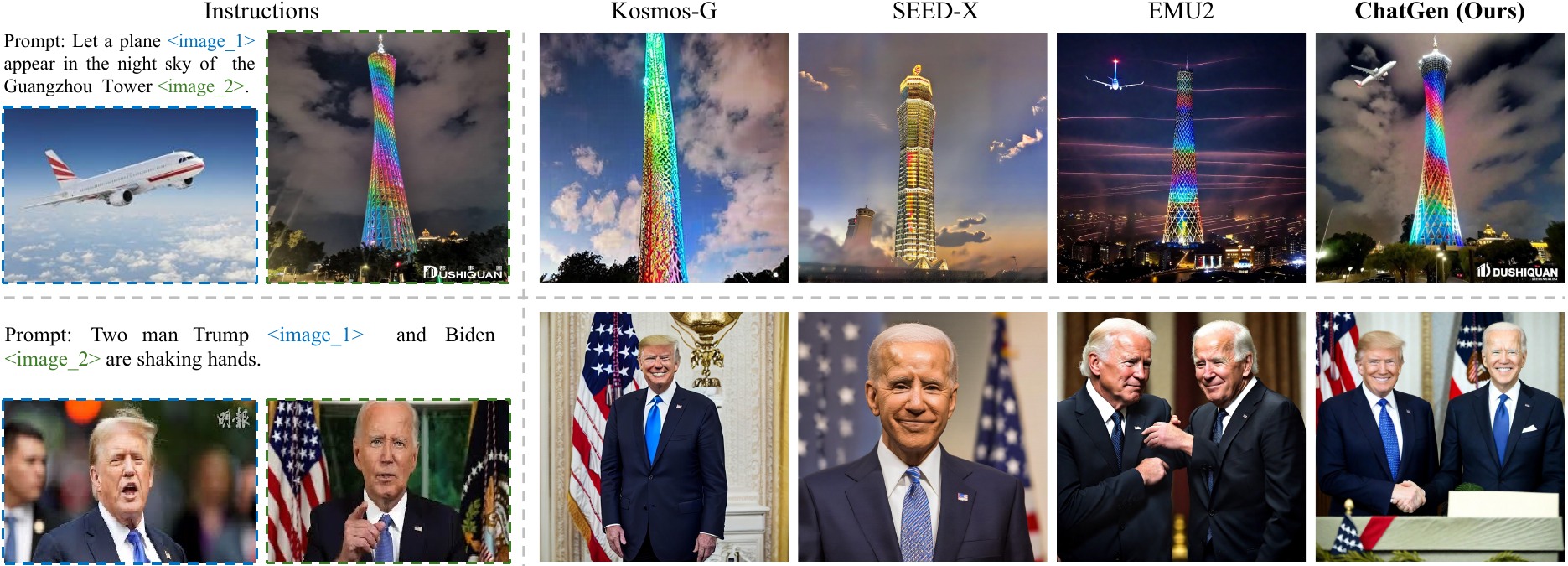
        }
        \vspace{-0.7cm}
        \caption{Comparison of instance identity consistency with state-of-the-art
        methods.}
        \label{fig:grounding_result1}
        \vspace{-0.5cm}
    \end{figure*}

    Dynamic instance identity consistency with reference images is crucial for
    practical applications, where users need to preserve specific instance details
    while allowing variations in other aspects. We evaluate our model's
    consistency through three perspectives: 1) qualitative comparisons with
    state-of-the-art methods (Fig.~\ref{fig:grounding_result1}), 2) quantitative
    evaluation on single subject-driven generation benchmarks (Table~\ref{tab:single_subject}),
    and performance on a new multi-character benchmark that better reflects real-world
    scenario (see Supplementary).

    \noindent
    \textbf{Comparative Analysis with SOTA Methods.} As shown in Figure~\ref{fig:grounding_result1},
    given reference images (leftmost column), we show how different methods perform
    when asked to modify specific attributes while maintaining subject identity.
    Previous methods either lose critical identity features or produce unnatural
    artifacts when attempting to change specific attributes. In contrast, our
    approach successfully preserves key identity characteristics while naturally
    incorporating the requested changes, demonstrating superior balance between consistency
    and variation. This improved performance can be attributed to our
    \ourdatasetshort data pipeline and enhanced visual encoder, more discussion is
    demonstrated in our ablation studies.


    \begin{table}[t]
        \centering
        \resizebox{0.95\linewidth}{!}{
        \begin{tabular}{@{}lccc@{}}
            \toprule \textbf{Methods}                            & \textbf{DINO} (↑) & \textbf{CLIP-I} (↑) & \textbf{CLIP-T} (↑) \\
            Re-Imagen~\cite{chen2022re}                          & 0.600             & 0.740               & 0.270               \\
            BLIP-Diffusion~\cite{li2024blip}                     & 0.594             & 0.779               & 0.300               \\
            \midrule SEED-X~\cite{ge2024seed}                    & 0.702             & 0.819               & 0.290               \\
            Kosmos-G~\cite{pan2023kosmos}                        & 0.694             & 0.847               & 0.287               \\
            Emu2-Gen~\cite{sun2024generative}                    & 0.766             & 0.850               & 0.287               \\
            OmniGen~\cite{xiao2024omnigenunifiedimagegeneration} & 0.801             & 0.847               & 0.301               \\
            \rowcolor[gray]{0.9} \textbf{\ours(Ours)}            & \textbf{0.823}    & \textbf{0.882}      & \textbf{0.302}      \\
            \bottomrule
        \end{tabular}
        }
        \caption{Quantitative comparison of zero-shot single-entity subject-driven
        generation on DreamBench, evaluation of instance consistency with identity
        preservation and natural variations.}
        \label{tab:single_subject}
    \end{table}

    \noindent
    \textbf{Single Subject-Driven Visual Generation.} Following the protocol in
    Kosmos-G~\cite{pan2023kosmos}, we evaluate \ours's subject-driven image generation
    capabilities on DreamBench~\cite{ruiz2023dreambooth}. For each prompt, we generate
    four images, totaling 3,000 images for a comprehensive evaluation. We use DINO~\cite{caron2021emerging}
    and CLIP-I~\cite{radford2021learning} to assess subject fidelity, and CLIP-T~\cite{radford2021learning}
    for text fidelity, in line with DreamBooth. Notably, \ours excels in subject
    fidelity, outperforming methods like BLIP-Diffusion and Kosmos-G on DINO and
    CLIP-I metrics.


    \subsection{Generation Diversity}
    \label{subsec:exp_diversity}

\begin{figure}[t]
        \centering
        \includegraphics[width=0.48\textwidth]{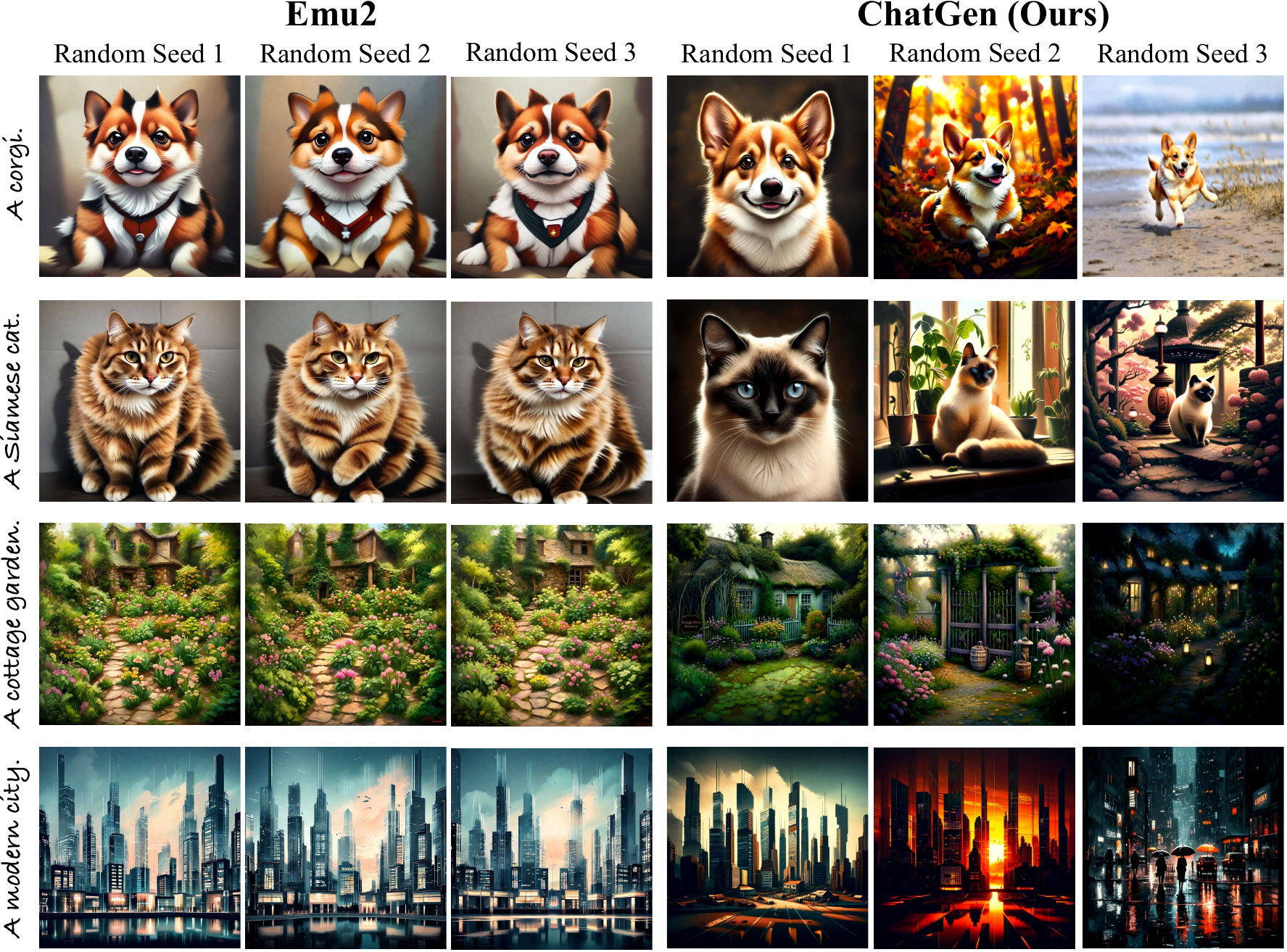}
        \vspace{-0.7cm}
        \caption{Diversity comparison of generated images with different random
        seeds. For each prompt, we show multiple generations from Emu-2 (left)
        and our method (right). Our method produces more diverse outputs while maintaining
        semantic consistency with the input prompts.}
        \label{fig:random_output}
        \vspace{-0cm}
    \end{figure}

    Generation diversity is another crucial capability of \ourtarget, as it enables
    users to explore various creative possibilities from a single prompt.
    However, existing methods often struggle with this aspect, producing nearly
    identical outputs despite different random seeds. As shown in Figure~\ref{fig:random_output},
    when given prompts like ``A corgi", ``A Siamese cat", ``A cottage garden",
    and ``A modern cityscape", EMU-2 generates highly similar images across
    different random seeds, limiting user choice. In contrast, our method produces
    diverse yet semantically consistent results for each prompt.

    \begin{table}[t]
        \centering
        \resizebox{0.28\textwidth}{!}{
        \begin{tabular}{lcc}
            \toprule Method                           & PSNR$_{d}$ $\downarrow$ & LPIPS$_{d}$ $\uparrow$ \\
            \midrule SEED-X~\cite{ge2024seed}         & 15.78                   & 0.2292                 \\
            Emu2-Gen~\cite{sun2024generative}         & 18.34                   & 0.2104                 \\
            \rowcolor[gray]{0.9} \textbf{\ours(Ours)} & \textbf{9.66}           & \textbf{0.6286}        \\
            \bottomrule
        \end{tabular}
        }
        \caption{Quantitative evaluation of generation diversity using PSNR$_{d}$
        and LPIPS$_{d}$ metrics. Lower PSNR$_{d}$ and higher LPIPS$_{d}$ values indicate
        greater diversity between samples.}
        \label{tab:diversity}
    \end{table}

    To quantitatively evaluate generation diversity, we compare samples generated
    with different random seeds using PSNR$_{d}$ and LPIPS$_{d}$ metrics (Table~\ref{tab:diversity};
    see supplementary material for detailed evaluation protocol). Our method achieves
    lower PSNR$_{d}$ and higher LPIPS$_{d}$ scores compared to SEED-X~\cite{ge2024seed}
    and Emu2~\cite{sun2024generative}, indicating greater diversity between
    generated samples. This enhanced diversity stems from our \ourrewriteshort,
    which introduces controlled randomness through sampling different prompt
    rewriting variants, allowing the model to explore diverse yet semantically consistent
    interpretations of the same input prompt.

    \subsection{Ablation Study}
    \label{subsec:ablation}

    To systematically evaluate our key design choices, we conduct comprehensive ablation
    studies focusing on three core aspects: instance identity consistency, generation
    diversity. Below, we analyze each component's contribution in detail.

\begin{figure}[t]
        \centering
        \includegraphics[width=0.48\textwidth]{
            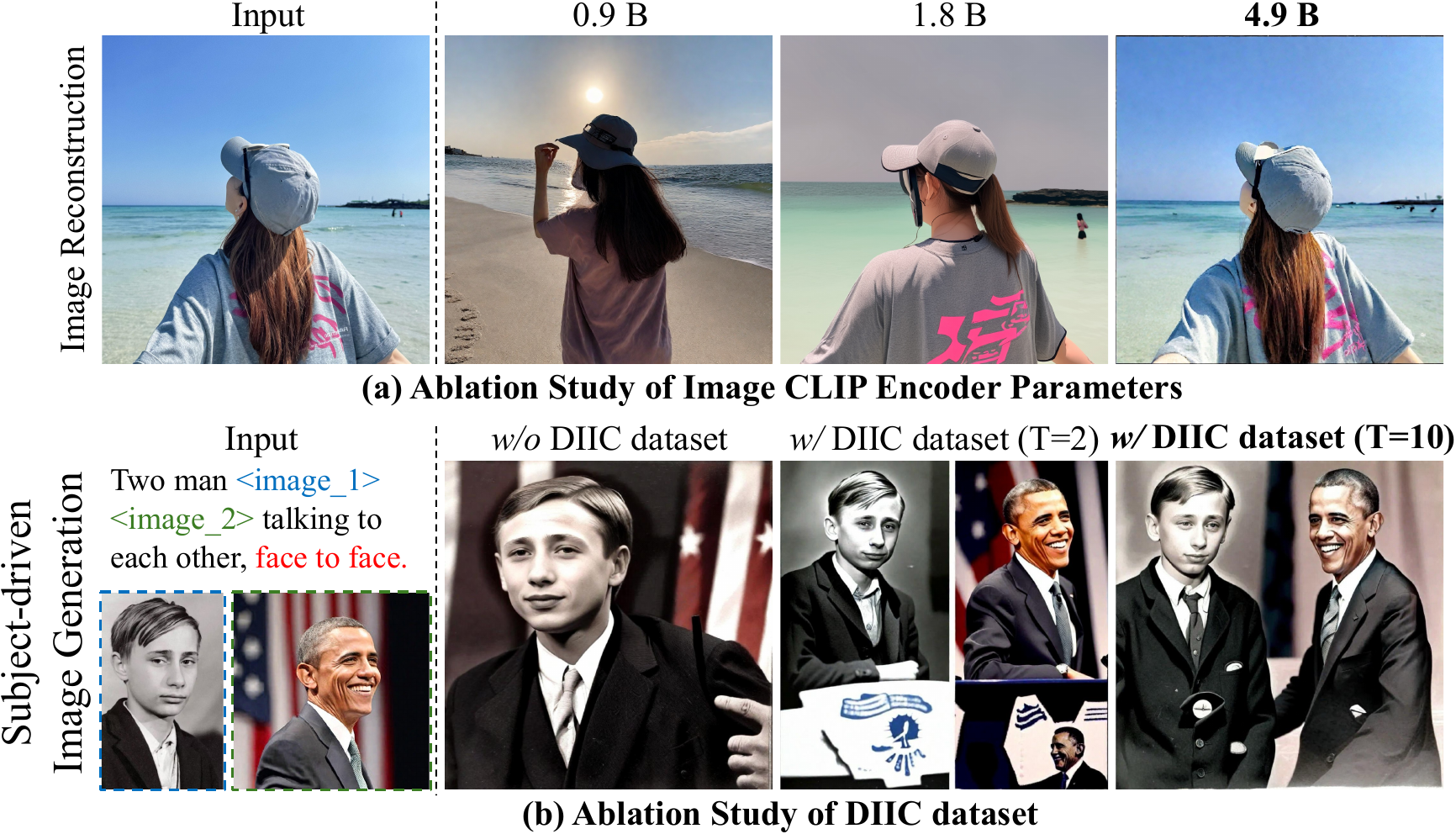
        }
        \vspace{-0.6cm}
        \caption{Ablation studies visualizing the impact of different components:
        (a) Effect of CLIP encoder scale on reconstruction quality; (b) Impact
        of \ourdatasetshort data pipeline and temporal sampling interval
        $t_{ref}$ on instance identity consistency and natural variations.}
        \label{fig:ablation_vis}
        \vspace{-0.4cm}
    \end{figure}

    \noindent
    \textbf{Enhanced Visual Encoder.} We first investigate how the scale of CLIP
    visual encoder affects reconstruction quality and instance identity
    consistency. Figure~\ref{fig:ablation_vis}(a) provides qualitative examples
    that demonstrate progressively better detail preservation and reconstruction
    quality with larger encoders. This improvement can be attributed to larger CLIP
    models' enhanced capability in extracting fine-grained visual features while
    maintaining a semantically meaningful latent space. The ability to accurately
    reconstruct reference images serves as a foundation for more complex instance
    identity consistency tasks - if a model struggles with basic reconstruction fidelity,
    it cannot be expected to maintain instance identity consistency in more
    challenging scenarios like attribute editing or subject-driven generation.


    \noindent
    \textbf{\ourdatasetshort Data-pipeline.} We analyze the effectiveness of our
    \ourdatasetshort data pipeline in balancing instance identity consistency.
    Quantitative results in Table~\ref{tab:ablation_tdic} show that removing
    \ourdatasetshort significantly degrades consistency metrics. The temporal sampling
    interval $t_{ref}$ plays a crucial role - a small interval leads to copy-paste
    behavior (high DINO but low CLIP-T scores), while a larger interval achieves
    better balance. Figure~\ref{fig:ablation_vis}(b) demonstrates these effects
    visually: the model without \ourdatasetshort fails at visual grounding,
    $t_{ref}$=2 produces copy-paste artifacts, while $t_{ref}$=25 successfully maintains
    key attributes while allowing natural variations.

    \begin{table}[t]
        \centering
        \begin{minipage}[t]{0.48\linewidth}
            \centering
            \scalebox{0.75}{
            \begin{tabular}{@{}lcc@{}}
                \toprule Setting              & DINO $\uparrow$ & CLIP-T $\uparrow$ \\
                \midrule w/o \ourdatasetshort & 0.684           & 0.300             \\
                $t_{ref}$=2                   & 0.835           & 0.264             \\
                $t_{ref}$=8                   & 0.831           & 0.272             \\
                $t_{ref}$=25                  & 0.823           & 0.302             \\
                $t_{ref}$=50                  & 0.801           & 0.302             \\
                \bottomrule
            \end{tabular}
            }
            \caption{Ablation of \ourdatasetshort dataset and temporal sampling
            interval $t_{ref}$ on DreamBench.}
            \label{tab:ablation_tdic}
        \end{minipage}
        \hfill
        \begin{minipage}[t]{0.48\linewidth}
            \centering
            \scalebox{0.75}{
            \begin{tabular}{@{}lcc@{}}
                \toprule Strategy  & PSNR$_{d}$ $\downarrow$ & CLIP-T $\uparrow$ \\
                \midrule w/o samp. & 19.88                   & 0.305             \\
                Pure samp.         & 7.34                    & 0.292             \\
                Top-P              & 8.44                    & 0.298             \\
                Temp               & 8.38                    & 0.301             \\
                Top-P+Temp         & 9.66                    & 0.308             \\
                \bottomrule
            \end{tabular}
            }
            \caption{Impact of sampling strategies on quality and diversity on
            COCO2014 dataset.}
            \label{tab:ablation_sampling}
        \end{minipage}
    \end{table}

    \noindent
    \textbf{\ourdatasetlong Mechanism.} We evaluate how our \ourrewriteshort strategy
    affects generation diversity. As shown in Table~\ref{tab:ablation_sampling},
    without this mechanism, the model produces nearly identical outputs across
    different random seeds. We then investigate various sampling strategies,
    including pure sampling, nucleus sampling (Top-P), temperature sampling, and
    their combination. Table~\ref{tab:ablation_sampling} shows that combining
    Top-P (p=0.9) with temperature sampling (t=0.8) achieves the best balance
    between generation original prompt following (CLIP-T) and diversity (PSNR$_{d}$).


    \section{Conclusion}

    In this work, we delve the unification of multimodal understanding and
    generation, landing as an interactive generation paradigm, \ieno, \ours.
    Compared to previous multimodal generation models, \ours exhibits superior capabilities
    in generating diverse outputs and maintaining consistency with instructions and
    reference images. This makes it particularly well-suited as a user-friendly
    design copilot. When user instructions are less detailed, \ours unleashes
    its creativity to produce diverse generation results, offering inspiration to
    the user. On the other hand, when users have more specific requirements,
    \ours adapts by refining its outputs based on the instructions and prior
    generations. During such refinements, it preserves consistency in the parts
    that the user is already satisfied with. Besides the unification modeling,
    we curate \ourdatasetshort, a large-scale dataset extracted from Internet videos
    and auto-labeled by advances foundation models to support learning to generate
    consistency-aware object dynamics. In addition, we further propose \ourrewriteshort,
    an effective mechanism to control the diversity of generation results. Extensive
    experiments demonstrate that the unified modeling of multimodal understanding
    and generation in \ours enables more controllable outputs, aligning better
    with user needs.

    { \small \bibliographystyle{ieeenat_fullname} \bibliography{main} }


    \clearpage

    \noindent
    \textbf{[Supplementary Material]}

    This supplementary material provides additional technical details (\S\ref{sec_supp:details}),
    extended experimental results (\S\ref{sec_supp:results}), and discusses
    limitations (\S\ref{sec_supp:limitation}) of \ours.

    \section{More Details about \ours}
    \label{sec_supp:details}

\begin{figure}[t!]
        \centering
        \includegraphics[width=0.7\linewidth]{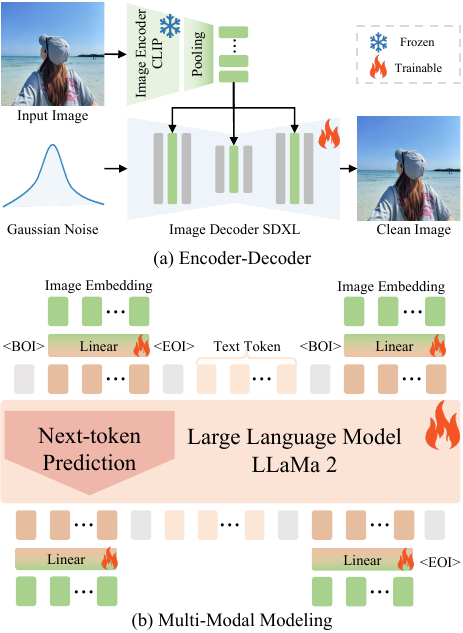}
        \caption{Detailed architecture of \ours.}
        \label{fig_supp:arch_detail}
    \end{figure}

\begin{figure}[ht]
        \centering
        \includegraphics[width=\linewidth]{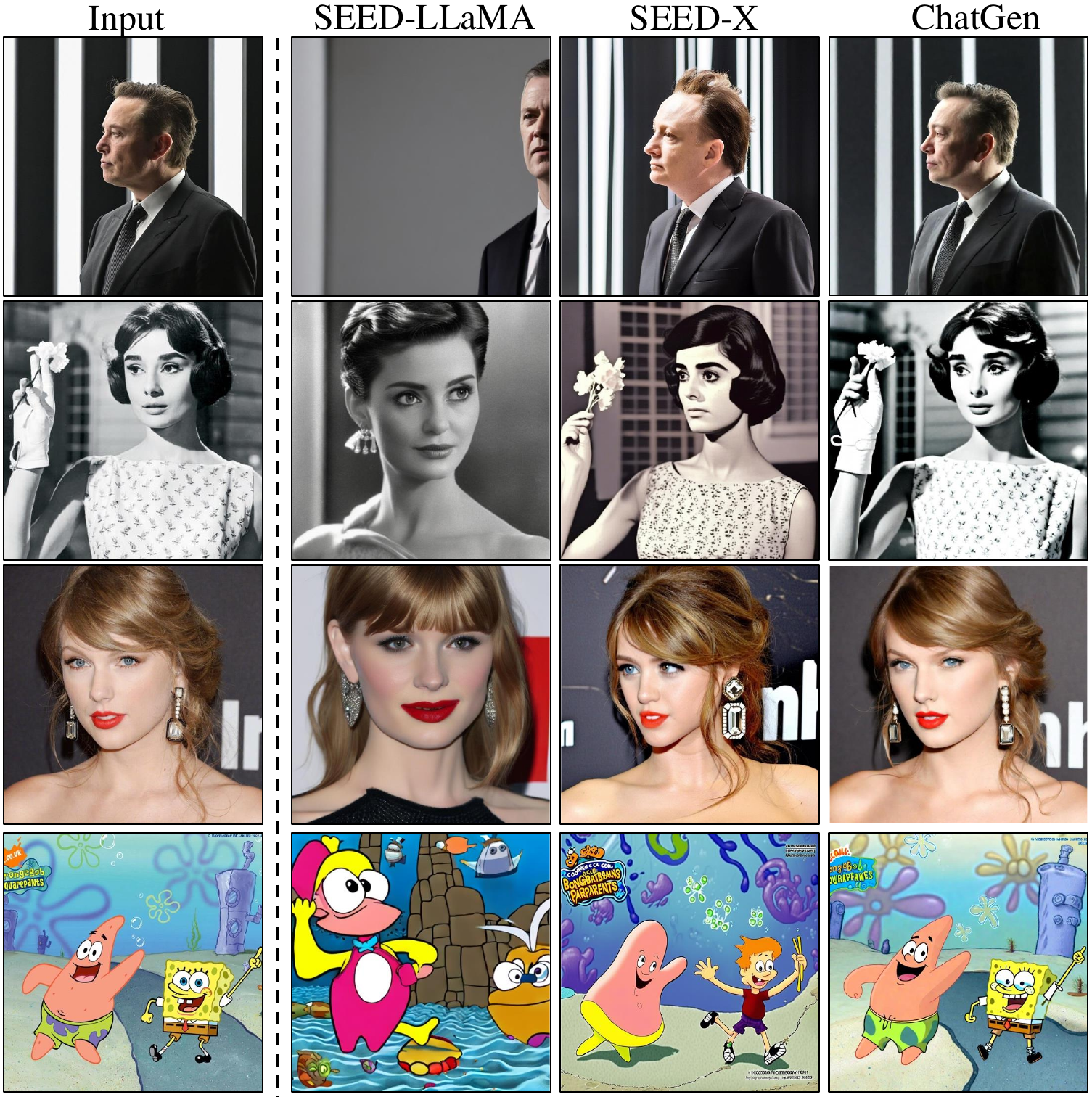}
        \caption{Qualitative comparison of reconstruction quality.}
        \label{fig_supp:recon_compare}
    \end{figure}

    \noindent
    \textbf{Visual Encoder-Decoder.} As shown in Fig.~\ref{fig_supp:arch_detail},
    unlike VAE-based approaches, we adopt the CLIP model as our image encoder to
    leverage its semantic extraction capabilities, enabling efficient text-visual
    joint modeling with significantly reduced training cost and data requirements
    (Table 1 in the main paper). However, CLIP encoders often struggle with preserving
    fine-grained visual details. As discussed in the main paper, we observe that
    larger CLIP models better maintain visual details while preserving semantic extraction.
    Based on this, we employ a pretrained EVA-CLIP~\cite{sun2023eva}(4.9B) as our
    image encoder. Through bicubic interpolation of position embeddings, we extend
    the encoder to process 448$\times$448 inputs instead of its original 224$\times$224
    resolution. The encoder outputs 16$\times$16$\times$1792 feature maps, which
    are pooled into a 64$\times$1792 sequence, preserving both semantic
    information and visual details. For the decoder, we fully fine-tune SDXL's
    UNet weights, using a learning rate of 5e-4 with cosine scheduling and classifier-free
    guidance training by randomly drop 10\% input image features. As shown in Figure~\ref{fig_supp:recon_compare},
    this configuration achieves superior reconstruction quality compared to
    existing methods.

\begin{figure}[ht!]
        \centering
        \includegraphics[width=0.8\linewidth]{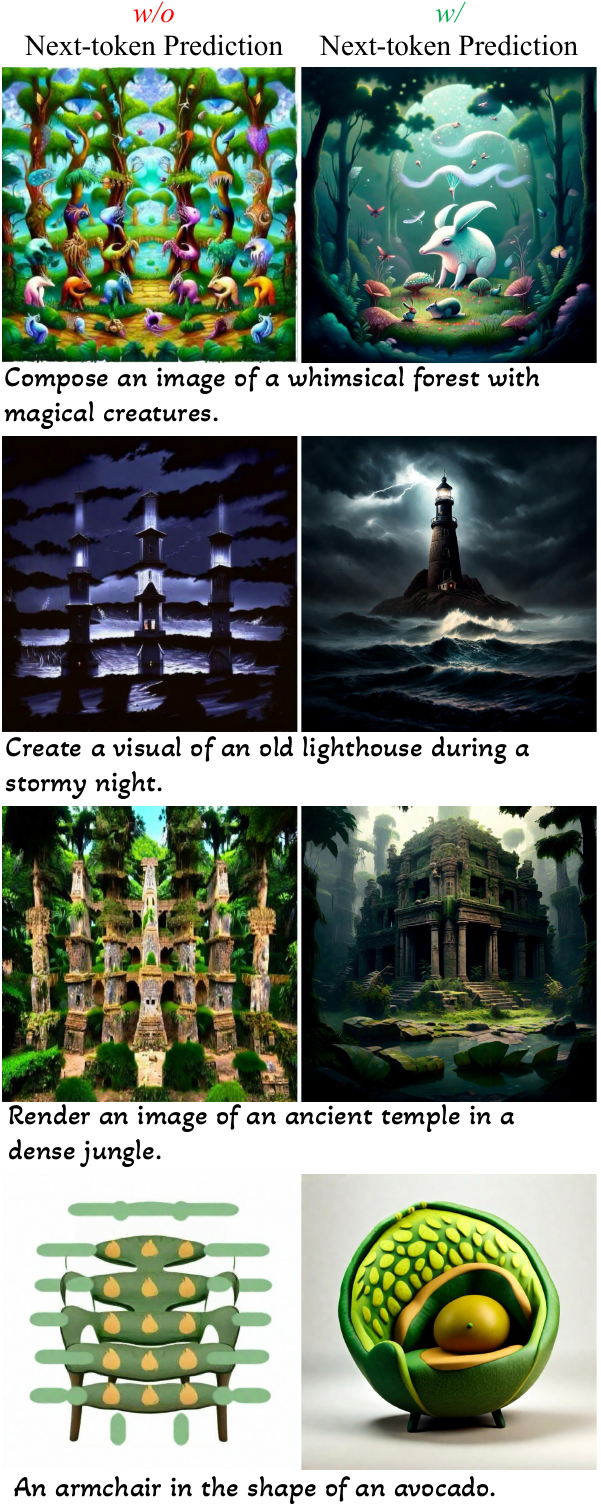}
        \caption{Visualization of feature modeling results. Left: parallel
        generation showing blocking artifacts. Right: our autoregressive
        generation producing more coherent visual features.}
        \label{fig_supp:multi_modal_feature_modeling}
    \end{figure}

    \noindent
    \textbf{Multi-modal Feature Modeling.} As shown in Fig.~\ref{fig_supp:arch_detail},
    we adopt an autoregressive approach for visual feature modeling. Unlike parallel
    generation methods~\cite{ge2023planting,ge2023making,ge2024seed} that simultaneously
    predict all visual features from fixed placeholder tokens (\egno $<$img1$>$
    to $<$img64$>$), our approach generates features sequentially with explicit
    dependencies:
    \begin{equation}
        P(x|c) = \prod_{i=1}^{64}P(x_{i}|x_{<i},c)
    \end{equation}
    \noindent
    This explicit modeling of inter-feature dependencies enables our model to better
    capture holistic visual structure. Each term $P(x_{i}|x_{i-1},...,x_{1},c)$
    leverages previously generated features as context, rather than generating features
    in isolation ($P(x_{i})$, $P(x_{i-1})$ ...). As shown in Figure~\ref{fig_supp:multi_modal_feature_modeling},
    the quality difference becomes more evident with a fully fine-tuned UNet
    decoder. This is because when UNet focuses purely on decoding, generation
    quality heavily depends on MLLM's visual feature modeling, the parallel approach
    (left) shows blocking artifacts due to independent feature generation, while
    our autoregressive approach (right) maintains coherence through contextual
    generation. While parallel visual modeling approaches~\cite{ge2023planting,ge2023making,ge2024seed}
    rely on SDXL's pretrained weights and inherent generation capability to compensate
    for weaker MLLM visual feature modeling, this dependency on the original
    SDXL decoder limits the MLLM's fine-grained control over generation and
    editing tasks, making it challenging to achieve a truly unified visual design
    copilot.

    \noindent
    \textbf{Dataset Details} Table~\ref{tab_supp:data_overview} presents a comprehensive
    overview of the diverse datasets used for training our model. Our training
    leverages two primary datasets: (1) \ourdatasetshort, containing 35M high-resolution
    frames with an average of 4.9 instances per frame and detailed captions (mean
    length 25.4 tokens); (2) Laion-COCO-Recaption, comprising 600M image-text
    pairs, each paired with both a concise caption (mean 10.2 tokens) and its expanded
    description (mean 79.6 tokens).

    \begin{figure*}[ht!]
        \centering
        \includegraphics[width=\linewidth]{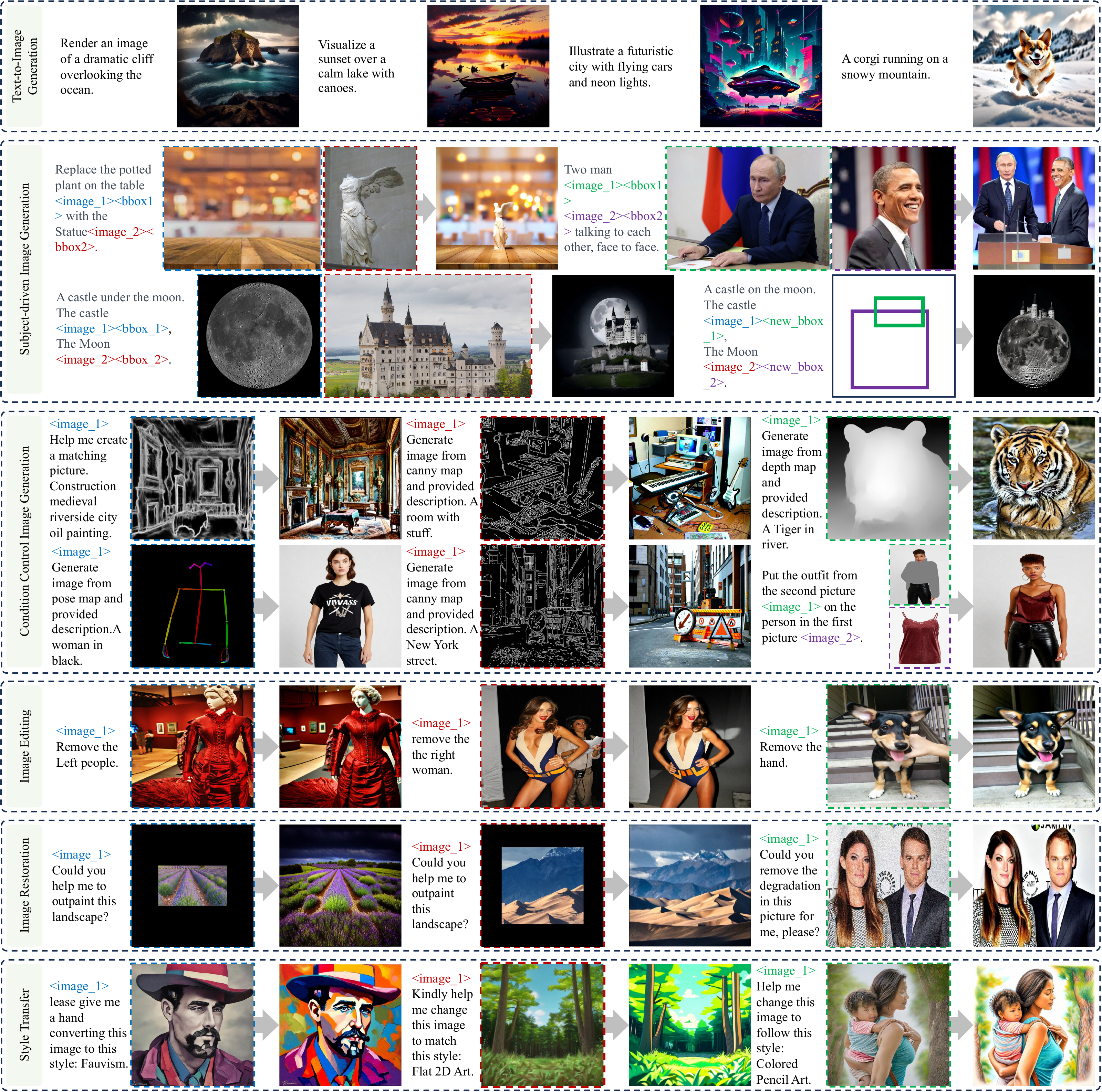}
        \caption{Extended case studies demonstrating \ours's diverse
        capabilities across multiple visual generation tasks.}
        \label{fig_supp:case_studies_extended}
    \end{figure*}

    \begin{table}[t]
        \centering
        \resizebox{\linewidth}{!}{
        \begin{tabular}{l|l}
            \toprule \textbf{Task}  & \textbf{Dataset}                                                                                                                                                   \\
            \midrule Reconstruction & \makecell[l]{Laion-COCO~\cite{laion2023coco}, Object365~\cite{shao2019objects365},\\ OpenImages~\cite{kuznetsova2020open}}                                         \\
            \hline
            Text2Image              & \makecell[l]{\textbf{Laion-COCO-Recaption}(Ours),\\ CapsFusion~\cite{yu2024capsfusion}, JourneyDB~\cite{sun2024journeydb}}                                         \\
            \hline
            Subject-Driven          & GrIT~\cite{peng2023kosmos}, \textbf{\ourdatasetshort}(Ours)                                                                                                        \\
            \hline
            Restoration             & \makecell[l]{Laion-COCO~\cite{laion2023coco}(Self-Aug),\\ MultiGen-20M~\cite{qin2023unicontrol}}                                                                   \\
            \hline
            Editing                 & \makecell[l]{SEED-Edit~\cite{ge2024seed}, MagicBrush~\cite{zhang2024magicbrush}}                                                                                   \\
            \hline
            Condition Gen           & MultiGen-20M~\cite{qin2023unicontrol}, HR-VITON~\cite{lee2022high}                                                                                                 \\
            \hline
            Style Transfer          & StyleBooth~\cite{han2024stylebooth}, MultiGen-20M~\cite{qin2023unicontrol}                                                                                         \\
            \hline
            Understanding           & \makecell[l]{\textbf{Laion-COCO-Recaption}(Ours), LLaVA-150K~\cite{liu2023visualinstructiontuning},\\ LLaVAR~\cite{zhang2023llavar}, ScienceQA~\cite{lu2022learn}} \\
            \bottomrule
        \end{tabular}
        }
        \caption{Overview of training datasets.}
        \label{tab_supp:data_overview}
    \end{table}



    \begin{table}[ht]
        \centering
        \setlength{\tabcolsep}{0.15cm}
        \renewcommand{\arraystretch}{1.2}
        \resizebox{\linewidth}{!}{
        \begin{tabular}{l|ccc}
            \toprule Configuration                         & Visual Decoding                        & Stage 1                                   & Stage 2            \\
            \midrule                                        
            Optimizer                                      & \multicolumn{3}{c}{AdamW}               \\
            Adam ($\beta_{1}$, $\beta_{2}$, $\varepsilon$) & $(0.9, 0.999, 10^{-8})$                & \multicolumn{2}{c}{$(0.9, 0.95, 10^{-6})$} \\
            Peak LR                                        & $5 \times 10^{-4}$                     & $5 \times 10^{-4}$                        & $1 \times 10^{-4}$ \\
            LR schedule                                    & \multicolumn{3}{c}{cosine decay}        \\
            Gradient clip                                  & 1.0                                    & \multicolumn{2}{c}{5.0}                    \\
            Training steps                                 & 5k                                     & 15k                                       & 5k                 \\
            Warmup steps                                   & \multicolumn{3}{c}{1000}                \\
            batch size                                     & 4096                                   & \multicolumn{2}{c}{2048}                   \\
            precision                                      & \multicolumn{3}{c}{$\mathtt{bfloat16}$} \\
            \bottomrule
        \end{tabular}
        }
        \caption{Training hyperparameters across different stages.}
    \end{table}

    \section{Additional Evaluation Results}
    \label{sec_supp:results}


    \noindent
    \textbf{Multi-Subject Generation Benchmark.} We construct a multi-subject generation
    benchmark using CelebA-HQ~\cite{zhu2022celebv} dataset, containing 2000 test
    cases with GPT-4 generated interaction prompts. Each case includes 2-3 reference
    faces. We evaluate using CLIP-T for text-image alignment, CLIP-I, DINO and
    face similarity\footnote{Using face\_recognition library (\url{https://github.com/ageitgey/face_recognition})}
    between reference and generated faces for identity preservation. As shown in
    Table~\ref{tab_supp:multi_subject}, \ours achieves superior performance across
    all metrics.

    \begin{table}[ht]
        \centering
        \resizebox{0.95\linewidth}{!}{
        \begin{tabular}{lcccc}
            \toprule Method      & DINO ($\uparrow$) & CLIP-I ($\uparrow$) & Face Sim. ($\uparrow$) & CLIP-T ($\uparrow$) \\
            \midrule Kosmos-G    & 0.583             & 0.712               & 19.1                   & 0.285               \\
            Emu2                 & 0.773             & 0.801               & 30.4                   & 0.294               \\
            SEED-X               & 0.664             & 0.709               & 20.8                   & 0.291               \\
            \textbf{\ours(Ours)} & \textbf{0.803}    & \textbf{0.845}      & \textbf{52.4}          & \textbf{0.294}      \\
            \bottomrule
        \end{tabular}
        }
        \caption{Performance comparison on multi-subject generation benchmark. Face
        Sim. denotes face similarity.}
        \label{tab_supp:multi_subject}
    \end{table}

    \noindent
    \textbf{Understanding Capabilities.} As shown in Table~\ref{tab_supp:understanding},
    while our primary focus is on unified visual generation for a design copilot,
    \ours still achieves superior understanding performance\footnote{All
    benchmarks are evaluated using VLMEvalKit (\url{https://github.com/open-compass/VLMEvalKit})}
    among unified models and maintains comparable results with understanding-only
    models across various visual understanding benchmarks.

    \begin{table*}
        [htbp]
        \centering
        \resizebox{\linewidth}{!}{
        \begin{tabular}{llcccccccccc}
            \toprule Type                                   & Models                                      & LLM Params & MMMU ($\uparrow$) & Hallusion ($\uparrow$) & MME ($\uparrow$) & MMStar ($\uparrow$) & MMT ($\uparrow$) & OCRBench ($\uparrow$) & ScienceQA ($\uparrow$) & MMVet ($\uparrow$) \\
            \midrule \multirow{6}{*}{\textit{Und.}}         & MiniGPT4~\cite{zhu2023minigpt}              & 7B         & 23.6              & 31.9                   & 1047.4           & 16.3                & 16.5             & 172                   & 39.6                   & 15.6               \\
                                                            & Kosmos-2~\cite{peng2023kosmos}              & 2B         & 23.7              & 19.8                   & 721.1            & 24.9                & 25.5             & 244                   & 32.7                   & 23.7               \\
                                                            & Idefics~\cite{idefics2023}                  & 9B         & 18.4              & 27.3                   & 1177.3           & 21.6                & 45.3             & 252                   & 53.5                   & 30.0               \\
                                                            & LLaVA~\cite{liu2023visualinstructiontuning} & 7B         & 34.1              & 21.6                   & 28.3             & 27.1                & 1075.5           & 269                   & 61.8                   & 28.3               \\
                                                            & Qwen-VL~\cite{bai2023qwen}                  & 7B         & 29.6              & 29.9                   & 482.7            & 32.5                & 42.9             & 127                   & 61.1                   & 13.0               \\
                                                            & Emu2-Chat~\cite{sun2024generative}          & 33B        & 40.7              & 29.5                   & 1678.0           & 40.7                & -                & 436                   & 68.2                   & 31.0               \\
            \midrule \multirow{5}{*}{\textit{Und. \& Gen.}} & Kosmos-G~\cite{pan2023kosmos}               & 1.9B       & 14.8              & 20.4                   & 104.3            & 18.4                & 18.3             & 109                   & 29.6                   & 11.3               \\
                                                            & Chameleon-7b~\cite{team2024chameleon}       & 7B         & 22.4              & 17.1                   & 202.7            & 31.1                & 23.9             & 5                     & 46.8                   & 8.3                \\
                                                            & Gemini-Nano-1~\cite{team2023gemini}         & 1.8B       & 26.3              & -                      & -                & -                   & -                & -                     & -                      & -                  \\
                                                            & LWM~\cite{liu2024world}                     & 7B         & -                 & -                      & -                & -                   & -                & -                     & -                      & 9.6                \\
                                                            & \textbf{\ours(Ours)}                        & 7B         & \textbf{26.6}     & \textbf{30.4}          & \textbf{447.4}   & \textbf{27.5}       & \textbf{28.4}    & \textbf{345}          & \textbf{63.1}          & \textbf{25.4}      \\
            \bottomrule
        \end{tabular}
        }
        \caption{Performance comparison on visual understanding benchmarks. Und.:
        Understanding-only models; Und. \& Gen.: Unified models with both understanding
        and generation capabilities.}
        \label{tab_supp:understanding}
    \end{table*}



    \noindent
    \textbf{Extended Case Studies.} Figure~\ref{fig_supp:case_studies_extended}
    presents additional examples showcasing \ours's capabilities across diverse
    tasks.

    \section{Limitations and Discussions}
    \label{sec_supp:limitation}

\begin{figure*}[ht!]
        \centering
        \includegraphics[width=\linewidth]{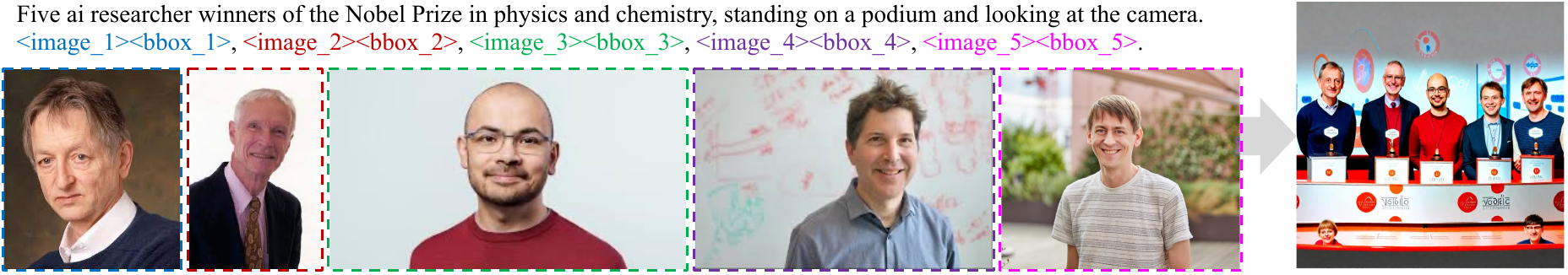}
        \caption{Failure cases with increasing number of reference images.}
        \label{fig_supp:limitation}
    \end{figure*}

    As shown in Figure~\ref{fig_supp:limitation}, our approach exhibits degraded
    instance-level consistency when handling multiple reference images. While performing
    well with 2-3 references, the identity preservation deteriorates as
    reference number increases.

\end{document}